\documentclass[runningheads]{llncs}

\usepackage{eccv}

\usepackage{eccvabbrv}

\usepackage{graphicx}
\usepackage{booktabs}

\usepackage[accsupp]{axessibility}  

\usepackage{amsfonts,amssymb}
\usepackage{hyperref}
\usepackage{wrapfig}

\usepackage{orcidlink}
\makeatletter
\newcommand{\printfnsymbol}[1]{%
\textsuperscript{\@fnsymbol{#1}}%
}
\begin{document}

\title{GPT-Connect: Interaction between Text-Driven Human Motion Generator and 3D Scenes in a Training-free Manner} 

\titlerunning{GPT-Connect}

\author{Haoxuan Qu\thanks{Both authors contributed equally to the work.}
\and
Ziyan Guo\printfnsymbol{1}
\and
Jun Liu\thanks{Corresponding Author}
}
\authorrunning{H. Qu et al.}
%
\institute{Singapore University of Technology and Design \\
\email{\{haoxuan\_qu, ziyan\_guo\}@mymail.sutd.edu.sg, jun\_liu@sutd.edu.sg}}

\maketitle

\begin{abstract}
Recently, while text-driven human motion generation has received massive research attention, most existing text-driven motion generators are generally only designed to generate motion sequences in a blank background. While this is the case, in practice, human beings naturally perform their motions in 3D scenes, rather than in a blank background. Considering this, we here aim to perform scene-aware text-drive motion generation instead. Yet, intuitively training a separate scene-aware motion generator in a supervised way can require a large amount of motion samples to be troublesomely collected and annotated in a large scale of different 3D scenes. To handle this task rather in a relatively convenient manner, in this paper, we propose a novel \textbf{GPT-connect} framework. In GPT-connect, we enable scene-aware motion sequences to be generated directly utilizing the existing blank-background human motion generator, via leveraging ChatGPT to connect the existing motion generator with the 3D scene in a totally training-free manner. Extensive experiments demonstrate the efficacy and generalizability of our proposed framework.
\end{abstract}

\section{Introduction}

Text-driven human motion generation aims to generate human motion sequences based on the given text prompts. 
It is relevant to various applications such as game development, film-making, and virtual reality experiences, and has received lots of research attention \cite{zhang2022motiondiffuse,tevet2022human,zhang2023t2m,wang2022humanise}. Among existing motion generation methods, most of them \cite{zhang2022motiondiffuse,tevet2022human,zhang2023t2m} focus on generating human motion sequences in a blank background. 
While this is important, in real life, human beings always interact with different 3D scenes, rather than performing their motions solely in a blank background. Considering this, another research problem that is important and worth exploring can be: how to generate the human motion sequence that can be based on the given text prompt, while also can interact with the given 3D scene properly. To tackle this problem, a potential solution that has been explored in the previous work \cite{wang2022humanise} is to train a specific scene-aware text-driven motion generator through a supervised-training mechanism. Specifically, in \cite{wang2022humanise}, given a training dataset that contains the ground-truth motion sequences in different 3D scenes, the architecture of the motion generator is first designed such that during its process of generating motion sequences, it can condition on both the 3D scene and the text prompt. After that, the designed generator is then optimized to minimize the difference between the generated motion sequences and the ground-truth ones over the training dataset. 

In this paper, we argue that, such a supervised-training mechanism can be sub-optimal in handling this scene-aware text-driven motion generation task.
This is because, based on such a mechanism, to optimize a high-quality and generalizable generator, the training dataset is generally demanded to be both large-scale and with enough diversity. 
However, especially for this scene-aware task, collecting and annotating such a dataset can be very troublesome. 
This is because, to be both large-scale and with enough diversity, such a dataset not only needs to contain a large number of ground-truth motion sequences over different text prompts, but also needs to properly interact these motion sequences with a large number of different 3D scenes. 
Existing datasets that can be used for this task are generally of limited diversity, for example, PROX \cite{hassan2019resolving} only involves 12 indoor 3D scenes, whereas HUMANISE \cite{wang2022humanise} is also an indoor dataset and generally only focuses on constructing its text prompts over 4 actions including walk, sit, stand up, and lie. 
Such limited diversity over samples in these datasets can strongly restrict the applicability and generalizability of the existing supervised-training-based method \cite{wang2022humanise}, in generating plausible human motions over different text prompts in arbitrary 3D scenes.

\begin{figure*}[t]
  \centering
  \includegraphics[width=\textwidth]{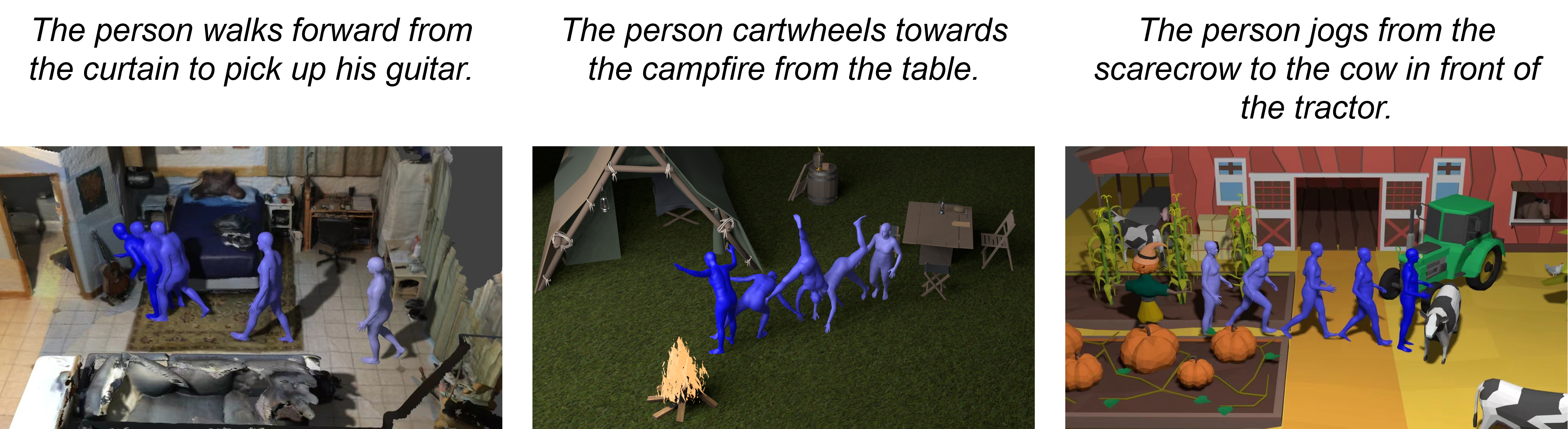}
\vspace{-0.2cm}
   \caption{Illustration of the scene-aware motion sequences generated by our GPT-Connect framework in different 3D scenes and based on different text prompts, in a totally training-free manner. As time passes, human meshes in the motion sequence are gradually changed from light to dark colors.}
   \label{fig:intro}
\vspace{-0.5cm}
\end{figure*}

Taking the above in mind, we here are wondering, rather than tackling this task relying on the supervised-training mechanism, from a novel perspective, if we can bring up a method that can handle this task instead in a totally training-free manner. To this end, we first notice that, the application of diffusion models has resulted in promising results in blank-background text-driven human motion generation \cite{zhang2022motiondiffuse,tevet2022human}. Considering this, in this paper, we aim to investigate, \textit{if we can find a connector, through which the off-the-shelf blank-background motion diffusion model can be automatically connected to the given 3D scene, and thus generate motion sequences that can properly interact with the given 3D scene without requiring any training.}
Note that if this can be achieved, we can handle the scene-aware text-driven human motion generation task in a pretty convenient manner, bypassing both the troublesome supervised-training process, and the demand of collecting and annotating a large-scale and diverse dataset that can be particularly tough for this task.

During our exploration, we find that, the off-the-shelf ChatGPT model \cite{ouyang2022training} may have the potential in playing this connector role. This is getting inspiration from the observation that, pre-trained over a tremendously large training corpus, ChatGPT has been found to contain rich implicit common-sense knowledge \cite{qu2023lmc}, and it has thus already played various different roles successfully in the many natural language processing tasks \cite{qin2023chatgpt,jiao2023chatgpt}. However, despite the success of ChatGPT in various natural language processing tasks, how to utilize ChatGPT to be the connector here to properly connect the 3D scene and the off-the-shelf motion diffusion model is still non-trivial owing to the following challenges: (1) While the training corpus of ChatGPT generally can contain many natural sentences describing different 3D scenes, a 3D scene by itself typically is not in a format that is ``friendly'' to ChatGPT. Thus, how to interpret a given 3D scene to ChatGPT, so that the implicit knowledge in ChatGPT can be effectively leveraged to understand this 3D scene, as well as how to interact with it, can be difficult. (2) Furthermore, even assuming certain designs have been equipped and ChatGPT has somehow understood the 3D scene, how ChatGPT can successfully pass its understanding as a guidance to the off-the-shelf motion diffusion model can also be challenging. 
To handle these challenges and enable ChatGPT to be the scene-generator-connector, in this paper, we propose a novel framework named \textbf{GPT-Connect}, which represents a new brand of method that can, for the first time, perform the scene-aware text-driven motion generation task in a totally training-free manner. We demonstrate the generalizability and flexibility of our framework in different 3D scenes in Fig.~\ref{fig:intro}, and outline our framework below.

Overall, in our framework, for enabling ChatGPT to be the connector between the 3D scene and the off-the-shelf motion diffusion model, we need to fulfill the two requirements as follows. 
(1) Firstly, given a 3D scene and a text prompt, our framework should ensure that ChatGPT can understand the 3D scene, and can correspondingly output ``useful information'' in guiding the motion sequence generated based on the text prompt to interact with the scene properly. 
(2) Moreover, given the ``useful information'' outputted by ChatGPT, our framework needs to further ensure that such information can be properly utilized by the motion diffusion model in guiding its motion generation process. 
In the following, before explaining how our framework is designed to meet the above two requirements, we first introduce the format of the above-mentioned ``useful information'' in our framework. Specifically, in our framework, we use (partial) skeleton sequences to be the ``useful information'' that ChatGPT passes to the motion diffusion model. This is because, firstly, with the notice that skeleton sequences can be regarded as a special combination of joint coordinates containing human priors, as can be seen in Sec.~\ref{sec:scene_channel} for more detail, when equipped with proper designs, ChatGPT has been found to hold the ability in outputting (partial) skeleton sequences. Meanwhile, skeleton sequences, as a simplified version of motion sequences, intuitively can demonstrate the motion diffusion model on how to interact with the scene.

After defining the ``useful information'' in our framework, below, we introduce how we fulfill the above two requirements, via incorporating our framework with the following two designs (channels).
Firstly, to properly interact ChatGPT with 3D scenes, in our framework, we introduce a Scene-GPT channel. Through this channel, the given 3D scene can be first interpreted in a format that is understandable by ChatGPT. Based on this (together with the text prompt given), under proper designs, ChatGPT can then output the ``useful information'' in a (partial) skeleton sequence format. Note that this outputted ``useful information'' can be regarded as implying a plausible way for the motion sequence to be generated based on the given text prompt, while also interacting with the 3D scene properly.
Secondly, we further incorporate our framework with a GPT-Generator channel. In this channel, we utilize the ``useful information'' outputted from ChatGPT to guide the generation process of the motion diffusion model, to generate plausible and scene-aware motion sequences. In summary, equipped with the above two channels, in our framework, we can enable ChatGPT to perform the scene-model-connector properly, and thus handle the scene-aware text-driven human motion generation task without requiring any training.

The contributions of our work are summarized as follows.
1) 
We propose GPT-Connect, a novel framework that for the first time, can handle the scene-aware text-driven human motion generation task in a totally training-free manner, via leveraging ChatGPT to be the intermediate connector between the 3D scene and the off-the-shelf motion diffusion model.
2) 
We introduce several designs in GPT-Connect to facilitate ChatGPT in better performing its scene-generator-connector role.
3) 
Without requiring any further training, GPT-Connect achieves state-of-the-art performance on the evaluated benchmarks, and can be flexibly used in different 3D scenes.

\section{Related Work}

\textbf{Text-driven Human Motion Generation.} Considering its wide range of applications, text-driven human motion generation has received a lot of research attention in the past years \cite{zhang2022motiondiffuse,tevet2022human,zhang2023t2m,wang2022humanise,ahuja2019language2pose,petrovich2022temos,guo2022generating,tevet2022motionclip,chen2023executing,karunratanakul2023optimizing,song2023loss,xie2023omnicontrol,shafir2023human,wang2023intercontrol,jiang2023motiongpt,zhang2023motiongpt,jing2023amd,yuan2023physdiff,dabral2023mofusion,guo2022tm2t,wei2023understanding,huang2023diffusion}. 
Before the application of diffusion models, different methods have proposed different ways to tackle this task. Among them, both T2M \cite{guo2022generating} and TEMOS \cite{petrovich2022temos} proposed to tackle this task via leveraging the variational autoencoder structure, whereas MotionCLIP \cite{tevet2022motionclip} proposed to seek help from CLIP's \cite{radford2021learning} shared text-image latent space. Later on, Tevet et al. \cite{tevet2022human} proposed to handle this task via the motion diffusion model, and as time passed, the application of the diffusion model has been gradually found an effective way to handle this task. More recently, Chen et al. \cite{chen2023executing} explored tackling this task in the latent space, and proposed the motion latent-based diffusion model, while certain other works enabled human beings to pass other types of conditions to the diffusion model as well besides texts, such as audios \cite{dabral2023mofusion} and trajectories \cite{karunratanakul2023guided,xie2023omnicontrol}. Alongside the exploration of the diffusion model in this task, recently, several works \cite{zhang2023motiongpt,jiang2023motiongpt} have also proposed to tackle this task via training a large transformer model. 

As an extension of the above works which generally focus on generating motion sequences more precisely in a blank background, recently, Wang et al. \cite{wang2022humanise} focused on the task of generating motion sequences that can be corresponding to the given text prompt, while at the same time can properly interact with the 3D scenes. Specifically, they proposed to tackle this scene-aware text-driven motion generation task via training a specific motion generation model through a supervised-training mechanism. Besides \cite{wang2022humanise} that leverages the supervised-training mechanism, there also exists another line of works \cite{juravsky2022padl,xiao2024unified} that explores training scene-aware text-driven motion generators through reinforcement learning. 

In this work, for the first time, we propose a framework that can perform the scene-aware text-driven motion generation task in a totally training-free manner. Different from the above existing motion generation methods, in our framework, we require neither motion samples to be annotated in the 3D scenes, nor human beings to provide guidance to the diffusion model. Instead, incorporated with proper designs, our GPT-Connect framework enables the off-the-shelf blank-background motion diffusion model to be automatically connected to the 3D scenes with the help of ChatGPT.

\noindent\textbf{Large Language Models.} Generally pre-trained over large corpora, recently, large language models such as ChatGPT \cite{ouyang2022training} have been found to contain rich common-sense knowledge \cite{qu2023lmc}. Considering this, in different tasks, large language models have been utilized to play various different roles \cite{qin2023chatgpt,jiao2023chatgpt,lian2023llm,shao2023prompting,qu2023lmc}, such as the role of language translator \cite{jiao2023chatgpt} and the role of question answerer \cite{shao2023prompting}. In this work, we design a novel framework, in which via equipped with proper designs, we enable the off-the-shelf ChatGPT model to play the connector role between the 3D scene and the off-the-shelf motion diffusion model.

\section{Proposed Method}

Given a 3D scene and a text prompt, the goal of scene-aware text-driven motion generation is to generate a human motion sequence that is based on both the text prompt and the 3D scene. To handle this task, as shown in the previous work \cite{wang2022humanise}, an intuitive method is to train a specific scene-aware text-driven motion generator through a supervised-training mechanism. Yet, to specifically optimize a high-quality and generalizable generator, such a mechanism typically demands collecting and annotating a large scale of different motion samples based on various text prompts in different 3D scenes, which can be very difficult. Considering this, in this work, we aim to bypass the tough demand of collecting and annotating motion samples in different 3D scenes, and instead handle this task more conveniently \textit{in a totally training-free manner}. To achieve this, we propose a new framework \textbf{GPT-Connect}, in which we propose to guide the off-the-shelf blank-background motion diffusion model to generate motion sequences that can properly interact with the given 3D scene (i.e., scene-aware motion sequences), via leveraging ChatGPT to be the intermediate connector between the 3D scene and the off-the-shelf diffusion model.

Specifically, denote the given 3D scene $S_{3D}$ and the given text prompt $t$. Our framework first incorporates a Scene-GPT channel. 
In this channel, ChatGPT is first guided to understand the 3D scene $S_{3D}$. 
After that, with the text prompt $t$ also passed to ChatGPT, based on both $t$ and $S_{3D}$, ChatGPT is instructed to output certain ``useful information'' $s[m_s]$, in the format of a partial skeleton sequence, where $s \in \mathbb{R}^{N \times J \times 3}$ represents the complete skeleton sequence with the same length as the motion sequence generated by the motion diffusion model, $N$ refers to the number of frames of $s$, and $J$ refers to the number of body joints in each frame of $s$. 
Besides, $m_s \in \{0, 1\}^{N \times J}$ represents the binary activation mask corresponding to $s$, where the [$n,j$]-th element of $m_s$ is 1 only if the coordinates of the $j$-th joint in the $n$-th frame of $s$ is outputted by ChatGPT through the Scene-GPT channel as part of the ``useful information''.
Intuitively, $s[m_s]$ roughly implies a way that the motion sequence generated based on the text prompt $t$ can interact with $S_{3D}$ properly. Considering this, after deriving $s[m_s]$, our framework then passes $s[m_s]$ to its GPT-Generator channel. Through this channel, $s[m_s]$, as ChatGPT's prediction and may thus be rough and inaccurate, can be properly leveraged to guide the off-the-shelf blank-background motion diffusion model to generate the motion sequence that is scene-aware. Below, we first describe the incorporated GPT-Generator channel, and then introduce the Scene-GPT channel.

\subsection{GPT-Generator Channel}
\label{sec:model_channel}

In this section, we first give a brief review of the (motion) diffusion model. We then introduce the GPT-Generator channel of our framework. 

\textbf{Revisiting diffusion models.} Analogous to non-equilibrium thermodynamic \cite{sohl2015deep}, the diffusion model generally consists of two parts, i.e., the forward process that iteratively diffuses the clean sample towards the noise, and the reverse process that gradually denoises the noise towards the clean sample. 
Note that among the two processes, during inference, we typically only need to perform the reverse process. Due to the effectiveness of the diffusion model, it has been applied in various vision areas \cite{ho2020denoising,meng2021sdedit,foo2023distribution}, such as image generation \cite{ho2020denoising} and image editing \cite{meng2021sdedit}. Especially among them, Tevet et al. \cite{tevet2022human} have applied diffusion models in the text-drive motion generation task to generate motion sequences in a blank background, and proposed the currently popularly-used motion diffusion model. Below, we focus on revisiting this motion diffusion model, which due to its effectiveness, is the off-the-shelf model that we utilize in our framework.

Specifically, in the motion diffusion model, the forward process aims to diffuse a clean motion sample $x_0$ iteratively towards a noise $x_K \sim \mathcal{N}(\textbf{0}, \textbf{I})$, in which the posterior distribution $q(x_k|x_{k-1})$ from $k = 1$ to $k = K$ is formulated as:
\begin{equation} \label{eq:revisiting_1}
\setlength{\abovedisplayskip}{3pt}
\setlength{\belowdisplayskip}{3pt}
\begin{aligned}
& q(x_k|x_{k-1}) = \mathcal{N}(x_k; \sqrt{\alpha_k}x_{k-1}, (1-\alpha_k)\textbf{I})
\end{aligned}
\end{equation}
where $\{\alpha_k \in (0, 1)\}^K_{k=1}$ is a set of hyperparameters, and $\alpha_k$ would gradually be closer to zero when $k$ becomes larger.

On the other hand, the reverse process of the motion diffusion model aims to iteratively denoise $x_K$ back into $x_0$. In specific, this process can be formulated as:
\begin{equation} \label{eq:revisiting_2}
\setlength{\abovedisplayskip}{3pt}
\setlength{\belowdisplayskip}{3pt}
\begin{aligned}
& p_{\omega}(x_{k-1}|x_{k}, t) = \mathcal{N}(x_{k-1}; \mu_k(\omega), (1-\alpha_k)\textbf{I}) \\
& \mu_k(\omega) = \frac{\sqrt{\overline{\alpha}_{k-1}}(1 - \alpha_k)}{1 - \overline{\alpha}_k} \hat{x}^{k}_0(x_k, k, t; \omega) + \frac{\sqrt{\alpha_k}(1-\overline{\alpha}_{k-1})}{1 - \overline{\alpha}_k} x_k \\
& \hat{x}_0^{k}(x_k, k, t; \omega) = f_{MDM}(x_k, k, t; \omega)
\end{aligned}
\end{equation}
where $t$ denotes the given text prompt, $\overline{\alpha}_k = \prod^k_{s=1} \alpha_s$, and $f_{MDM}$ denotes the motion diffusion model with its parameter denoted as $\omega$. Note that in each step of the reverse process, the motion diffusion model $f_{MDM}$ predicts the final clean motion instead of the current step's noise. In Eq.~\ref{eq:revisiting_2}, we denote this predicted final clean motion as $\hat{x}^{k}_0(x_k, k, t; \omega)$. In the rest of this work, for simplicity, we call this prediction $\hat{x}^{k}_0$ and omit items in the bracket.

\textbf{The big picture of this channel.} 
In the GPT-Generator channel, besides the text prompt $t$ and the off-the-shelf text-drive motion diffusion model $f_{MDM}$, we are also given $s[m_s]$ as the channel's input. Recall that derived from the Scene-GPT channel, $s[m_s]$ can roughly imply a way by which the motion sequence generated based on $t$ can interact with $S_{3D}$ properly. 
Given these inputs, besides basing the reverse process on $t$, in this channel, we aim to additionally condition the reverse process of the motion diffusion model on $s[m_s]$.
By doing so, we can guide the motion sequence $x_0$ finally generated from the reverse diffusion process to interact with the 3D scene $S_{3D}$ properly. 

To achieve the above, for the reverse diffusion process as a step-to-step process gradually denoising from $x_K$ to $x_0$, at every step $k$ denoising from $x_k$ to $x_{k-1}$, a naive way can be to directly align the output $x_{k-1}$ of the current step towards $s[m_s]$. By doing so, through step-to-step alignments, ideally, we hope that $x_0$ as the output of the final step (i.e., when $k=1$) can fit $s[m_s]$ well. 
Yet, practically, this naive way can suffer from the following two problems and thus lead to suboptimal results: 
(1) In the reverse diffusion process, at every step $k$, its output $x_{k-1}$ belongs to a unique distribution $X_{k-1}$ with a unique noisy level that is different from other steps. Thus, unless at the last step when $k = 1$, $X_{k-1} \neq X_0$. Yet, at the same time, $s[m_s]$ can be regarded as a partial ``clean'' skeleton sequence. In other words, while the distribution of $s[m_s]$ can be regarded as close to $X_0$, it is not necessary for $s[m_s]$ to be also close to other $X_{k-1} \neq X_0$. Considering the above, unless at the last step (i.e., $k = 1$ and $x_{k-1} \in X_{k-1} = X_0$), directly and forcefully aligning $x_{k-1}$ towards $s[m_s]$ can pull $x_{k-1}$ away from its desired $X_{k-1}$ distribution space and thus interrupt the reverse process of the motion diffusion model. 
(2) Moreover, recall that $s[m_s]$ is derived from ChatGPT's output and it thus can contain inaccuracies. Thus, overly aligning $x_{k-1}$ towards $s[m_s]$ throughout every step of the reverse process can suffer the final motion sequence $x_0$ from the inaccuracies in $s[m_s]$, which is also undesirable.
To tackle the above two problems, here in our GPT-Generator channel, we condition the reverse process of the motion diffusion model on $s[m_s]$ leveraging a new strategy, in which two modifications are made as follows over the above naive way of directly aligning $x_{k-1}$ towards $s[m_s]$ throughout the reverse process.

\textbf{Modification 1: aligning $\hat{x}^k_0$ towards $s[m_s]$ instead.} Firstly, in our GPT-Generator channel, at step $k$ of the reverse process, rather than directly aligning $x_{k-1}$ as the output of the current step towards $s[m_s]$, we instead first align $\hat{x}^k_0$ as an intermediate value of the current step towards $s[m_s]$, and then derive $x_{k-1}$ from the aligned version of $\hat{x}^k_0$ (i.e., $\widetilde{x}^k_0$). 
Recall that $\hat{x}^k_0$ represents the diffusion model's prediction of the final clean motion at step $k$ of its reverse process. Considering this, by performing alignment in this way, across different steps of the reverse process, the alignments are then consistently performed between $s[m_s]$ and elements in the distribution space $X_0$, and the reverse diffusion process would not suffer from the problem of pulling $x_{k-1}$ away from $X_{k-1}$.

Specifically, to achieve the alignment between $s[m_s]$ and $\hat{x}^k_0$, we first define the gap between them as:
\begin{equation} \label{eq:diffusion_1}
\setlength{\abovedisplayskip}{3pt}
\setlength{\belowdisplayskip}{3pt}
\begin{aligned}
g_k(s[m_s], \hat{x}^k_0) = ||s[m_s] - f_{m-s}(\hat{x}^k_0)[m_s]||_2
\end{aligned}
\end{equation}
where $f_{m-s}(\cdot)$ represents the function that projects a motion sequence to its corresponding skeleton sequence. Note that after such a motion-to-skeleton projection, $f_{m-s}(\hat{x}^k_0)$ then shares the same shape as $s$, i.e., $f_{m-s}(\hat{x}^k_0) \in \mathbb{R}^{N \times J \times 3}$. Thus, as shown in Eq.~\ref{eq:diffusion_1}, utilizing $f_{m-s}(\cdot)$, we can formulate the gap between $s[m_s]$ and $\hat{x}^k_0$ directly as the L2 distance between $s[m_s]$ and $f_{m\_s}(\hat{x}^k_0)[m_s]$. After formulating the gap between $s[m_s]$ and $\hat{x}^k_0$, in step $k$ of the reverse process, we then align $\hat{x}^k_0$ towards $s[m_s]$ and acquire $\widetilde{x}_0^k$ as the aligned version of $\hat{x}^k_0$:
\begin{equation} \label{eq:diffusion_2}
\setlength{\abovedisplayskip}{3pt}
\setlength{\belowdisplayskip}{3pt}
\begin{aligned}
& \widetilde{x}_0^k = \hat{x}^k_0 - \lambda \nabla_{x_k} g_k(s[m_s], \hat{x}^k_0)
\end{aligned}
\end{equation}
where $\lambda$ is a coefficient hyperparameter and $\nabla_{x_k} g_k(s[m_s], \hat{x}^k_0)$ is the gradient calculated from $g_k(s[m_s], \hat{x}^k_0)$. $x_{k-1}$ can then be finally acquired from $\widetilde{x}_0^k$ as:
\begin{equation} \label{eq:diffusion_3}
\setlength{\abovedisplayskip}{3pt}
\setlength{\belowdisplayskip}{3pt}
\begin{aligned}
& p_{\omega}(x_{k-1}|x_{k}, t) = \mathcal{N}(x_{k-1}; \mu_k(\omega), (1-\alpha_k)\textbf{I}) \\
& \mu_k(\omega) = \frac{\sqrt{\overline{\alpha}_{k-1}}(1 - \alpha_k)}{1 - \overline{\alpha}_k} \widetilde{x}^k_0 + \frac{\sqrt{\alpha_k}(1-\overline{\alpha}_{k-1})}{1 - \overline{\alpha}_k} x_k
\end{aligned}
\end{equation}
Note that via deriving $x_{k-1}$ in this way from $\widetilde{x}_0^k$, intuitively, we can indirectly condition $x_{k-1}$ on the useful information $s[m_s]$ through $\widetilde{x}_0^k$, while at the same time not pulling $x_{k-1}$ away from $X_{k-1}$. We also demonstrate the efficacy of this indirect alignment design empirically (as shown in Tab.~\ref{Tab:ablation_study_1}).

\textbf{Modification 2: selectively deactivating the alignment process in certain steps.} Additionally, we notice that $s[m_s]$, while being a very useful prior knowledge for interacting with the 3D scene properly, is only the prediction made by ChatGPT but not the ``ground-truth'' guidance. Thus, $s[m_s]$ may sometimes inevitably be non-natural or contain inaccuracies, especially considering that ChatGPT by itself is not a professional human skeleton expert. Taking this in mind, in our GPT-Generator channel, to keep the motion sequence generated to be realistic, while taking $s[m_s]$ as the guidance, we don't want to overly guide the motion generation process of the motion diffusion model with $s[m_s]$. Instead, at certain reverse diffusion steps when the gap $g_k$ defined in Eq.~\ref{eq:diffusion_1} between $s[m_s]$ and the current diffusion step is already small, we deactivate the alignment process and perform motion generation solely relying on the rich motion-related knowledge of the off-the-shelf motion diffusion model. Specifically, to implement this modification, each time we use the motion diffusion model to generate a motion sequence, at the first reverse step denoising from $X_K$ to $X_{K-1}$, we first store the corresponding gap $g_K$ measured by Eq.~\ref{eq:diffusion_1}. After that, in every step $k$ of the following diffusion steps, we measure $x_{k-1}$ from $\widetilde{x}_0^k$ as the aligned version of $\hat{x}_0^k$ only when $\frac{g_k}{g_K} > \xi$, where $\xi$ is a hyperparameter, otherwise, we measure $x_{k-1}$ directly from $\hat{x}_0^k$. 

After making the above two modifications, conditioning on $s[m_s]$ besides $t$, the reverse process of the motion diffusion model in our GPT-Generator channel is finally formulated as: 
\begin{equation} \label{eq:diffusion_4}
\setlength{\abovedisplayskip}{3pt}
\setlength{\belowdisplayskip}{3pt}
\begin{aligned}
& p_{\omega}(x_{k-1}|x_{k}, t) = \mathcal{N}(x_{k-1}; \mu_k(\omega), (1-\alpha_k)\textbf{I}) \\
& \mu_k(\omega) = 
    \begin{cases}
    \frac{\sqrt{\overline{\alpha}_{k-1}}(1 - \alpha_k)}{1 - \overline{\alpha}_k} \widetilde{x}^k_0 + \frac{\sqrt{\alpha_k}(1-\overline{\alpha}_{k-1})}{1 - \overline{\alpha}_k} x_k,  & \text{if } \frac{g_k}{g_K} > \xi \\
    \frac{\sqrt{\overline{\alpha}_{k-1}}(1 - \alpha_k)}{1 - \overline{\alpha}_k} \hat{x}^k_0 + \frac{\sqrt{\alpha_k}(1-\overline{\alpha}_{k-1})}{1 - \overline{\alpha}_k} x_k,  & \text{if } \frac{g_k}{g_K} \leq \xi
    \end{cases}    
\end{aligned}
\end{equation}
Note that, the motion sequence $x_0$ finally generated from the model diffusion model is of a fixed length $N$. Yet, in practice, over different 3D scenes and based on different text prompts, the scene-aware motion sequence that is plausible and is implied by $s[m_s]$ can have varying length. Considering this, rather than directly utilizing $x_0$ as the output of the current GPT-Generator channel, we propose to instead use the sub-sequence of $x_0$ that is properly guided by $s[m_s]$ as the channel's final output. Specifically, denoting $n_{start}$ and $n_{end}$ respectively the first and last frame in the binary mask $m_s$ that contains activated body joints, after deriving $x_0$, we then further clip $x_0^c$ from $x_0$ as $x_0^c = x_0[n_{start}:n_{end}]$, and take $x_0^c$ as the final output of this channel. By doing so, with the help of $s[m_s]$, we can finally leverage the off-the-shelf blank-background motion diffusion model, to generate the motion sequences that can properly interact with the 3D scene $S_{3D}$.

\subsection{Scene-GPT Channel}
\label{sec:scene_channel}

In the Scene-GPT channel of our framework, since $S_{3D}$ by itself is not understandable to ChatGPT, as a preparation, we first interpret (describe) $S_{3D}$ in a format that ChatGPT can understand. After that, given both the text prompt $t$ and this understandable description of $S_{3D}$, we instruct ChatGPT to output the ``useful information'' $s[m_s]$, which is then used in the GPT-Generator channel of our framework to guide the motion diffusion model as discussed in Sec.~\ref{sec:model_channel}.

\textbf{Understanding $S_{3D}$.} At the start of this Scene-GPT channel, we first aim to find a format that can not only describe $S_{3D}$, but also is understandable to ChatGPT. Specifically, we notice that, while $S_{3D}$ can be described from different perspectives, the key thing that can affect its interaction with the motion sequence is its layout, rather than the texture or appearance of the objects in the scene. Taking this in mind, we further observe that, ChatGPT has seen a large number of programming language scripts (e.g., CSS scripts) describing the layout of different web pages during its pre-training stage, while in such scripts, the layout of different objects on the web page is commonly described as a collection of their types and bounding boxes \cite{feng2024layoutgpt}. Inspired by this, here in this channel, we propose to also interpret $S_{3D}$ to ChatGPT as a collection of object types and bounding boxes in this 3D scene.

\begin{wrapfigure}[16]{r}{0.45\textwidth}
\vspace{-0.6cm}
\centering
\includegraphics[width=0.45\textwidth]{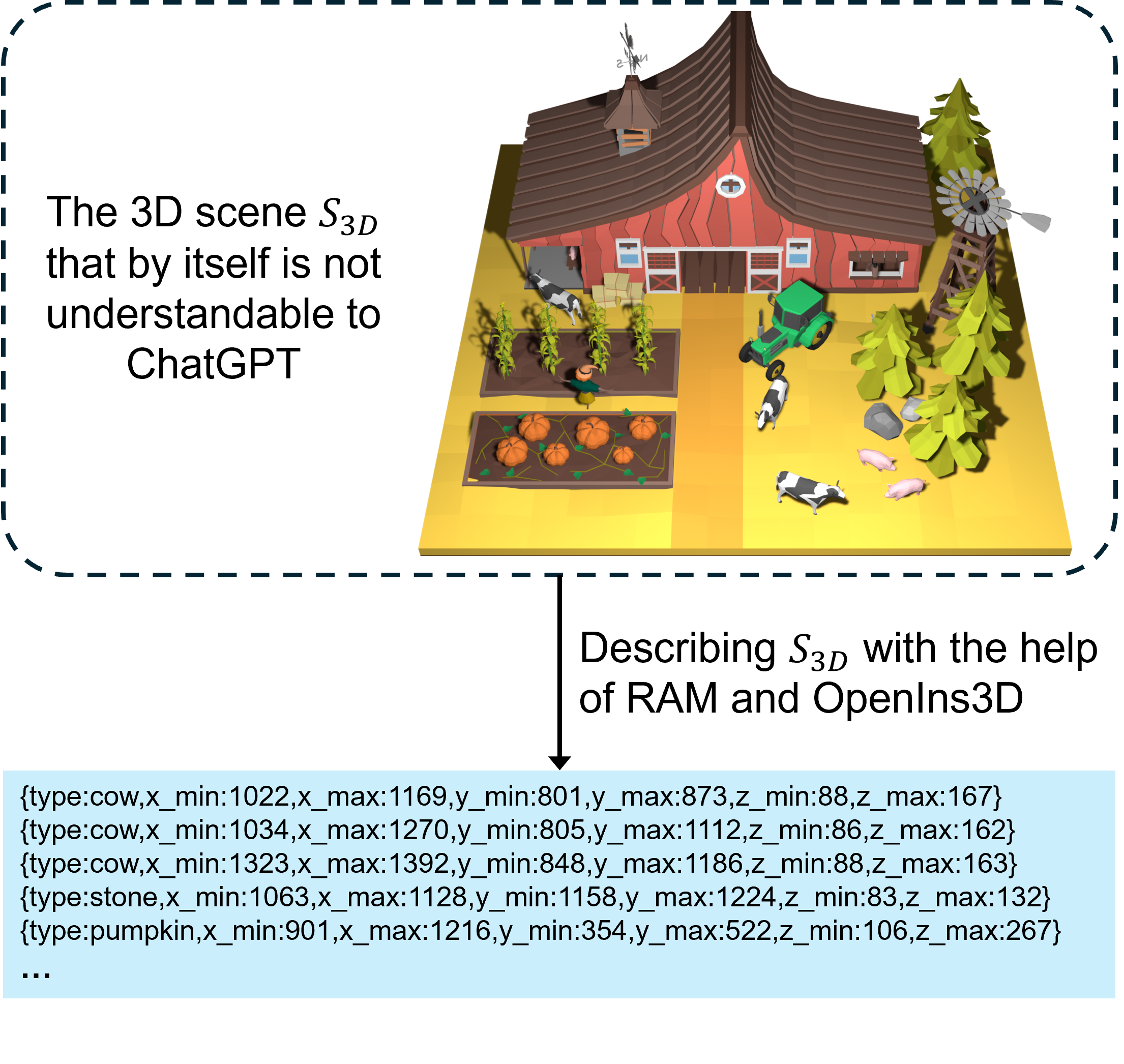}
\vspace{-0.8cm}
\caption{Illustration of the process of describing $S_{3D}$ in a format that is understandable to ChatGPT.}
\label{fig:scene}
\end{wrapfigure}
Specifically, to interpret $S_{3D}$ in this way, we perform the following 3 steps. \textbf{(1) Recognition.} Given $S_{3D}$, to bound each object in it, if we don't have any prior knowledge over $S_{3D}$, we first need to know what are the types of objects that can exist in this scene. To achieve this, with the notice that directly performing object recognition over a 3D scene can be difficult, here, we first render the 3D scene and take $M$ photos over it from $M$ different camera poses. After that, we leverage the off-the-shelf open-world recognition model Recognize Anything (RAM) \cite{zhang2023recognize}, to recognize the list of objects that exist in each of the taken photos. Finally, the unified set of the $M$ object lists derived from the $M$ photos is then regarded as all potential objects that can appear in $S_{3D}$ (more details on this recognition step are also provided in the supplementary). \textbf{(2) Bounding boxes derivation.} Once we get the list of objects that can potentially exist in $S_{3D}$ either from step (1) or from prior knowledge, we then pass both $S_{3D}$ and this list to an off-the-shelf 3D open-vocabulary instance segmenter to derive the type and the 3D bounding box over each instance in $S_{3D}$. Here in our framework, we use OpenIns3D \cite{huang2023openins3d} as this off-the-shelf segmenter. \textbf{(3) Passing to ChatGPT.} Finally, with types and 3D bound boxes over objects in $S_{3D}$ acquired, as shown in Fig.~\ref{fig:scene}, we can then interpret the layout of $S_{3D}$ to ChatGPT in a format that ChatGPT can understand, in which we finally represent the 3D bounding box of each object in the scene with a list of six values (i.e., x\_min, x\_max, y\_min, y\_max, z\_min, and z\_max).

\textbf{Deriving $s[m_s]$ from ChatGPT.} After describing $S_{3D}$ in a ChatGPT-understandable way as above, based on both the text prompt $t$ and the description of $S_{3D}$, we then guide ChatGPT to output the ``useful information'' $s[m_s]$, in the format of a partial skeleton sequence. Note that, ChatGPT naturally can have the potential in becoming a skeleton sequence generator and generating the ``useful information'' $s[m_s]$ properly. This is because, on the one hand, pre-trained over a tremendously large corpus that typically involves very rich descriptions of human skeletons and human motions \cite{brown2020language}, ChatGPT can contain very rich implicit knowledge \cite{ouyang2022training} of human skeleton structure and human behaviors; on the other hand, for skeleton sequences that essentially are combinations and concatenations of body joint coordinates, ChatGPT that has been pre-trained on massive programming data \cite{feng2024layoutgpt} has naturally seen a large number of numbers and coordinates during its pre-training stage. Yet, recall that ChatGPT typically has not been specifically pre-trained over skeleton sequences. Thus, directly asking ChatGPT to generate skeleton sequences that can properly interact with the scene can still be a very tough task for it. Considering this, to facilitate ChatGPT in understanding our demand and tacking this tough task, in the instruction we design to guide ChatGPT in generating the partial skeleton sequence, besides specifying the task goal, the input format, and the expected output format to ChatGPT, we also involve a step-by-step guidance mechanism.

Specifically, we get motivated by the existing work \cite{wei2022chain} that, large language models can better understand and tackle a certain task if they can first divide the task into several simpler sub-tasks and then handle each sub-task at a time in a step-by-step manner. Considering this, in our instruction passed to ChatGPT, we also divide the tough task of generating $s[m_s]$ first into several simpler sub-tasks, and guide ChatGPT to tackle them step-by-step. Specifically, we find that one effective way to achieve this can be to guide ChatGPT to tackle the following four sub-tasks: (1) ``First, please locate the target object mentioned in the text prompt by the provided bounding boxes.''; (2) ``Second, you should reason and plan for the motion trajectory including the starting point and the end point.''; (3) ``Third, you must reason about the reasonable initial orientation of the person.''; (4) ``Fourth, you must determine how many frames are in the motion according to the text prompt and the trajectory.''. Empirically, we observe that, after step-by-step guided by the above four sub-tasks, ChatGPT as a powerful off-the-shelf model can understand its given tough task, and generate $s[m_s]$ that can match both $t$ and $S_{3D}$ well. 
We also demonstrate both examples of $s[m_s]$ outputted by ChatGPT, and the complete instruction we pass to ChatGPT in the supplementary.

\subsection{Overall Inference Process}

Above we describe the two channels of our GPT-Connect framework. Here, we summarize the overall inference process of our framework. Specifically, given a 3D scene $S_{3D} $ and a text prompt $t$, in our framework, we first pass them to the Scene-GPT channel of our framework to get the ``useful information'' $s[m_s]$. After that, we pass both $s[m_s]$ and $t$ to the GPT-Generator channel of our framework, to derive the final output $x_0^c$ of our framework. 
Note that the whole inference process of our framework can be automatically performed with a script.

\section{Experiments}

To evaluate the efficacy of our proposed framework, in this section, we first evaluate our method quantitatively on the HUMANISE dataset \cite{wang2022humanise}. Moreover, we also compare our framework with the existing training-based method qualitatively on both the indoor 3D scenes in HUMANISE, and on other 3D scenes (e.g., outdoor 3D scenes) outside HUMANISE.

\begin{figure*}[t]
  \centering
  \includegraphics[width=\textwidth]{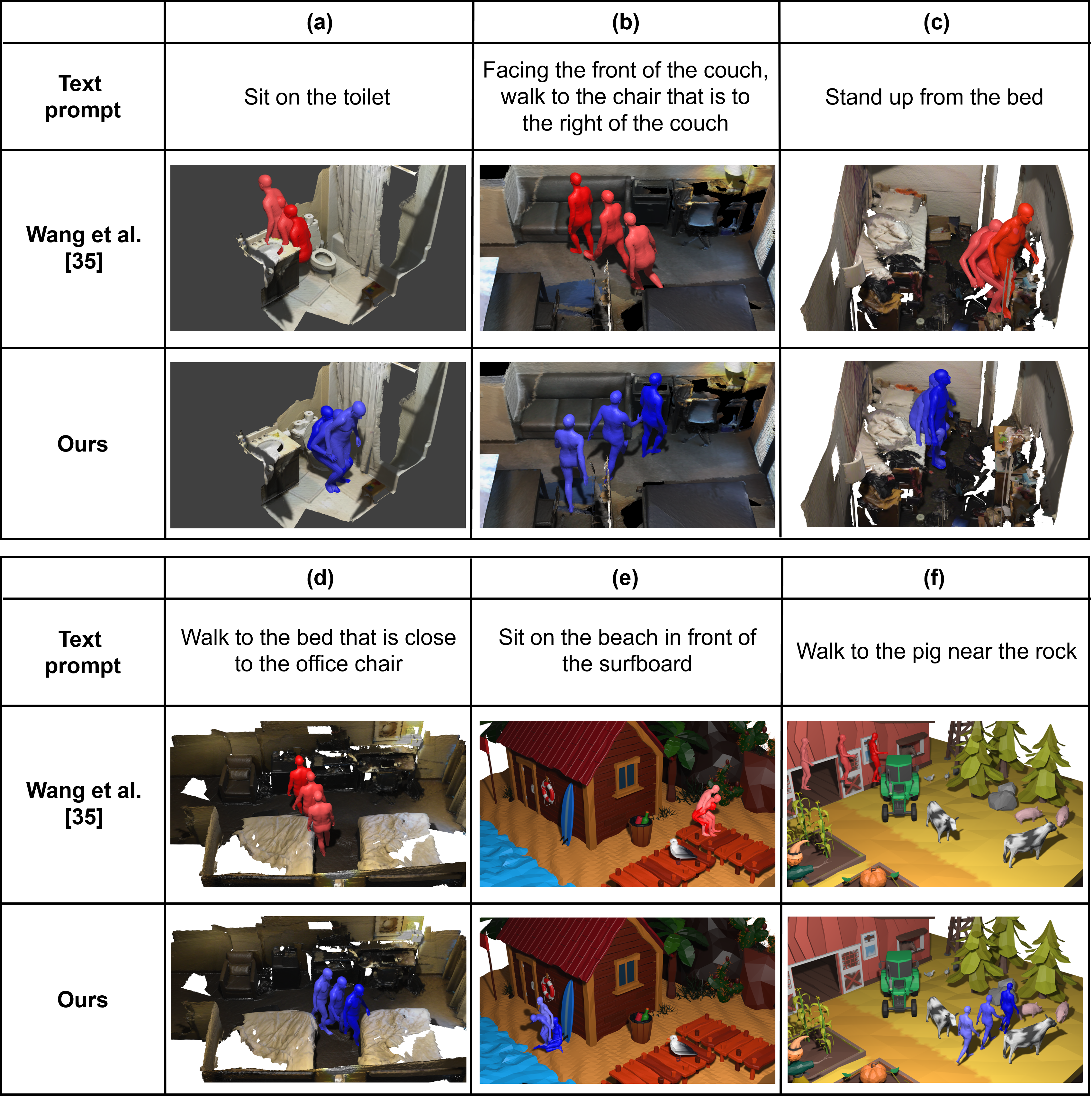}
\vspace{-0.5cm}
\caption{Qualitative results of our framework. (a-d) are on the 3D scenes in HUMANISE, while (e-f) are on outdoor 3D scenes outside HUMANISE. Light to dark colors of the human meshes denote time. More qualitative results are in the supplementary.}
   \label{fig:qualitative}
\vspace{-0.4cm}
\end{figure*}
\subsection{Dataset and Evaluation Metrics}

\textbf{HUMANISE} \cite{wang2022humanise} is a human motion dataset that contains 19.6k human motion sequences in 643 different 3D scenes. On this dataset, besides performing evaluation on the whole testing set, Wang et al.  \cite{wang2022humanise} also particularly perform further comparison between their method and two alternative baselines on a subset of the testing set that focuses on different text prompts describing the ``walking'' action. Following \cite{wang2022humanise}, we here evaluate our method on both the whole testing set and this ``walking'' subset in our experiments.

\noindent\textbf{Evaluation metrics.} To evaluate the quality of the text-driven motion sequences generated in different 3D scenes, on HUMANISE, following \cite{wang2022humanise}, we first adopt the following three metrics: the quality score, the action score, and the body-to-goal distance. 
Among them, to measure the quality score and the action score (each has its range from 1-5 and the higher the better), following \cite{wang2022humanise}, we perform a human perceptual study w.r.t. both the overall quality and the action-semantic accuracy of the generated motion sequences. While for the body-to-goal distance, it is calculated as the shortest distance (in meters) between the core object defined in the given text prompt, and the human motion sequence generated in the scene. To better measure the physical plausibility of the generated motion sequences in the 3D scene, besides the above three metrics, following \cite{wang2021synthesizing,wang2022towards}, we also involve two additional metrics: the non-collision score
and the contact score (each has its range from 0-1 and the higher the better). 
We also provide more details about the above five metrics in the supplementary.

\subsection{Implementation Details}

We conduct our experiments on RTX 3090 GPUs. For ChatGPT, we use GPT-4. Besides, for the motion diffusion model \cite{tevet2022human}, we use the ``humanml-encoder-512'' version of it downloaded directly from its official GitHub repo \cite{mdm}. Besides, we set the coefficient hyperparameter $\lambda$ used in Eq.~\ref{eq:diffusion_2} to be 3, the threshold $\xi$ used in Eq.~\ref{eq:diffusion_4} to be 0.2, and the number of photos $M$ that are taken over $S_{3D}$ from different camera poses to be 16. 

\begin{table}[t]
\caption{Results on the whole testing set of the HUMANISE dataset.}
\vspace{-0.2cm}
\centering
\resizebox{\textwidth}{!}
{\small
\setlength{\tabcolsep}{2pt}
\begin{tabular}{c|c|c|c|c|c} \hline
Method & Quality score \textuparrow & Action score \textuparrow & Body-to-goal distance \textdownarrow & Non-collision score \textuparrow & Contact score \textuparrow \\ \hline
Wang et al. \cite{wang2022humanise} & 2.57 & 3.59 & 1.01 & \textbf{1.00} & 0.73 \\
Ours  & \textbf{2.83} & \textbf{3.67} & \textbf{0.87} & \textbf{1.00} & \textbf{0.80} \\ \hline
\end{tabular}}
\vspace{-0.2cm}
\label{Tab:overall}
\end{table}

\begin{table}[t]
\caption{Results on the ``walking'' subset of the testing set of the HUMANISE dataset.}
\vspace{-0.2cm}
\centering
\resizebox{\textwidth}{!}
{\small
\setlength{\tabcolsep}{2pt}
\begin{tabular}{c|c|c|c|c|c} \hline
Method & Quality score \textuparrow & Action score \textuparrow & Body-to-goal distance \textdownarrow & Non-collision score \textuparrow & Contact score \textuparrow \\ \hline
Baseline A & 2.88 & 3.80 & 1.50 & - & - \\
Baseline B & 2.80 & 3.75 & 1.44 & - & - \\
Wang et al. \cite{wang2022humanise} & 2.91 & 3.88 & 1.37 & \textbf{1.00} & 0.69 \\
Ours  & \textbf{3.04} & \textbf{4.01} & \textbf{0.96} & \textbf{1.00} & \textbf{0.76} \\ \hline
\end{tabular}}
\vspace{-0.4cm}
\label{Tab:subset}
\end{table}

\subsection{Main Results}

\noindent\textbf{Quantitative results on the whole testing set of HUMANISE.} As shown in Tab.~\ref{Tab:overall}, compared to the existing method \cite{wang2022humanise}, in a totally training-free manner, our framework achieves superior performance across different evaluation metrics. This shows the effectiveness of our framework. 

\noindent\textbf{Quantitative results on the ``walking'' subset of HUMANISE.} In Tab.~\ref{Tab:subset}, following the evaluation setting of \cite{wang2022humanise}, we also perform further comparison on the ``walking'' subset of the testing set, among our framework, the method proposed in \cite{wang2022humanise}, and the two baselines (baseline A \& baseline B) formulated in \cite{wang2022humanise}. Intuitively, these two baselines differ from the method proposed in \cite{wang2022humanise} from the perspective that they design their scene-aware motion generators to be with different architectures (more details about these two baselines are in the supplementary). As shown in Tab.~\ref{Tab:subset}, our training-free framework consistently outperforms those existing training-based methods with different motion generator architectures, further demonstrating the superiority of our framework.

\noindent\textbf{Qualitative results.} Besides comparing our framework with the existing method \cite{wang2022humanise} quantitatively, in Fig.~\ref{fig:qualitative}, we also show qualitative results of our framework, over both the 3D scenes in HUMANISE, and the 3D scenes outside HUMANISE (collected from Sketchfab instead). As shown, across different 3D scenes and based on different text prompts, our framework can consistently generate plausible motion sequences. At the same time, our framework can also consistently interact with the 3D scene more properly than \cite{wang2022humanise}. This further shows our framework's efficacy and generalizability.

\subsection{Ablation Studies}

\noindent\textbf{Impact of the guidance strategy in the GPT-Generator channel.} In the GPT-Generator channel of our framework, to condition the reverse diffusion process on the ``useful information'' $s[m_s]$ effectively, we make two modifications over the naive way of directly aligning $x_{k-1}$ towards $s[m_s]$ in every step of the reverse diffusion process. To evaluate the efficacy of these modifications, we test three variants. In the first variant (\textit{w/o both modifications}), we directly condition the reverse diffusion process on $s[m_s]$ following the above naive way. In the second variant (\textit{w/o modification 1}), in those diffusion steps where the alignment is not deactivated by modification 2, we perform the alignment directly between $x_{k-1}$ and $s[m_s]$, but not between $\hat{x}_0^k$ and $s[m_s]$. Moreover, in the third variant (\textit{w/o modification 2}), rather than deactivating alignment in certain steps, we perform alignment between $\hat{x}_0^k$ and $s[m_s]$ in every step of the reverse process.
As shown in Tab.~\ref{Tab:ablation_study_1}, our framework outperforms all the three variants. This shows the effectiveness of both modifications we make over the guidance strategy in the GPT-Generator channel. \textbf{More ablation studies such as experiments w.r.t. the Scene-GPT channel and w.r.t. hyperparameters are in the supplementary.}

\begin{table}[h]
\vspace{-0.4cm}
\caption{Evaluation on the guidance strategy incorporated in the GPT-Generator channel on the whole testing set of HUMANISE.}
\vspace{-0.3cm}
\centering
\resizebox{\textwidth}{!}
{\small
\setlength{\tabcolsep}{2pt}
\begin{tabular}{c|c|c|c|c|c} \hline
Method & Quality score \textuparrow & Action score \textuparrow & Body-to-goal distance \textdownarrow & Non-collision score \textuparrow & Contact score \textuparrow \\ \hline\
w/o both modifications  & 2.22 & 3.06 & 1.27 & 0.99 & 0.68 \\
w/o modification 1  & 2.29 & 3.15 & 1.19 & 0.99 & 0.71 \\
w/o modification 2  & 2.75 & 3.52 & 0.94 & 1.00 & 0.77 \\
GPT-Connect  & 2.83 & 3.67 & 0.87 & 1.00 & 0.80 \\ \hline
\end{tabular}}
\vspace{-0.7cm}
\label{Tab:ablation_study_1}
\end{table}

\section{Conclusion}

In this paper, we have proposed a novel scene-aware text-driven motion generation framework GPT-Connect. In this framework, from a novel perspective via utilizing ChatGPT to be the intermediate connector between the 3D scene and the off-the-shelf motion diffusion model, we enable scene-aware motion sequences to be generated in a totally training-free manner. Specifically, we design both a Scene-GPT channel and a GPT-Generator channel in our framework to enable ChatGPT to perform its scene-generator-connector role well. Without requiring any training, our framework generalizes well to different indoor and outdoor 3D scenes, and consistently outperforms the previous training-based method.


%
%
\bibliographystyle{splncs04}
\bibliography{main}
\end{document}